\documentclass[letterpaper, 10pt, conference]{ieeeconf}
\pdfoutput=1
\IEEEoverridecommandlockouts
\usepackage[utf8]{inputenc}
\usepackage[T1]{fontenc}
\usepackage{algorithm}
\usepackage[noend]{algpseudocode}
\usepackage{amssymb}
\usepackage{amsmath}
\usepackage{array}
\usepackage{balance}
\usepackage{graphics}
\usepackage{epsfig}
\usepackage{mathptmx}
\usepackage{times}
\usepackage{color}
\usepackage{balance}

\newcommand{\hide}[1]{---HIDDEN---}
\renewcommand{\hide}[1]{#1} 

\DeclareMathAlphabet{\mathcal}{OMS}{cmsy}{m}{n}
\DeclareMathOperator*{\argmin}{arg\,min}

\newcolumntype{C}[1]{>{\centering\let\newline\\\arraybackslash\hspace{0pt}}m{#1}}

\title{\LARGE \bf
Active and Transfer Learning of Grasps by Sampling from Demonstration
}
\author{\hide{Philipp Zech and Justus Piater}%
\thanks{The research leading to these results has received funding from the \hide{European Community's Seventh Framework
        Programme FP7/2007--2013 (Specific Programme Cooperation, Theme 3, Information and Communication Technologies) under
        grant agreement no.~610532, SQUIRREL}.}%
\thanks{\hide{Philipp Zech and Justus Piater are with the
        Faculty of Mathematics, Computer Science and Physics,
        University of Innsbruck, 6020 Innsbruck, Austria
        \texttt{philipp.zech@uibk.ac.at} and \texttt{justus.piater@uibk.ac.at}}}%
}

\begin{document}

\maketitle
\thispagestyle{empty}
\pagestyle{empty}

\begin{abstract}
We guess humans start acquiring grasping skills as early as at the infant
stage by virtue of two key processes. First, infants attempt
to learn grasps for known objects by imitating humans. Secondly, knowledge acquired
during this process is reused in learning to grasp novel objects. We argue that
these processes of active and transfer learning boil down to a random search of
grasps on an object, suitably biased by prior experience. In this paper we introduce
active learning of grasps for known objects as well as transfer learning of grasps
for novel objects grounded on kernel adaptive, mode-hopping Markov Chain Monte Carlo.
Our experiments show promising applicability of our proposed learning methods.
\end{abstract}

\section{Introduction}\label{sec:1}

Efficiently learning successful robotic grasps is one of the key challenges
to solve for successfully exploiting robots for complex tasks. Considering
existing research, grasp learning methods can be grouped into analytic and
empirical (or data-driven) methods~\cite{bohg2014,sahbani2012}. Balasubramanian~\cite{bala2012}
showed that empirical grasp learning grounded upon Programming by Demonstration
(PbD) can achieve results superior to planner based, analytic methods.

PbD is a rather simple learning concept constructed from the idea of a robot
observing a human demonstrator to then autonomously learn manipulation skills
from its observations. Generally, these methods rely on recording hand
trajectories. These trajectories then are taken as a basis for either recognizing object
and hand shapes (obviously supported by vision), analytic computation of contact
points of successful grasps, or a combination of both to learn grasps~\cite{bohg2014}.
In this paper, we propose an alternate approach in that we sidestep the reliance
on hand trajectories. Instead, we only require a few user demonstrated grasps as
6D gripper poses. From these, we then learn new grasps by sampling gripper
poses relative to a canonical object pose. This ultimately results in a grasp
learning method that requires no object specific knowledge.

Treating a grasp as a 6D pose unlocks two key advantages compared to
shape-based and analytic methods. First, learned grasps are readily applicable
to known objects by just mapping the 6D gripper pose from a canonical object pose
to the actual object pose. This requires no further knowledge than
the actual object pose. Secondly, acquired grasping skills are easily transferred
to novel, as of yet unseen objects, by suitably biasing the learning process.
This is by virtue of objects that are similar in \emph{shape} and \emph{size}
usually have similar grasp affordances. Conversely, shape-based or analytic
approaches would require either reconstruction of a shape or computation of
new contact points which may easily fail due to clutter, improper segmentation,
or missing object information.

Metropolis-Hastings~\cite{hastings1970} is a popular Markov-Chain Monte Carlo
(MCMC) sampler that establishes a Markov chain on a state space $\mathcal{X}$
(e.g., the grasp parameter space) where the stationary distribution of the Markov
chain is the target probability density $\pi(x)$ sought-after. By iteratively
drawing samples $x_{i}$ from a proposal distribution $q(x|y)$ one can finally
approximate $\pi(x)$. We propose the application of kernel adaptive, mode-hopping
MCMC (Section~\ref{sec:3}) for (i) active learning of grasps for known objects
 and (ii) transfer learning for acquiring grasps for novel objects
to learn an object's grasp density $\pi(x)$ by sampling.

In this work we first introduce active learning of grasps for known objects by
combining MCMC Kameleon~\cite{sejdinovic2014} and Generalized Darting Monte Carlo
(GDMC)~\cite{smini2011} (Section~\ref{sec:4}). This requires both a rough sketch
of the shape of $\pi$ for the former and an initial set of modes (i.e., a set of
demonstrated grasps) of $\pi$ for the latter. Given this rough sketch MCMC
Kameleon then learns an approximation of $\pi$, while GDMC nudges the proposal
generating process to elliptical regions around modes of $\pi$ for efficient
mixing between modes. Secondly, we present transfer learning of grasps for novel
objects similar in \emph{shape} and \emph{size} to already learned objects
(Section~\ref{sec:5}). This primarily capitalizes on MCMC Kameleon's learning
behavior during a burn-in phase that allows learning of $\pi$ for a novel object
(e.g., a soup plate) by approximating it with the Markov chain of a similar
object (e.g., a plate). Additionally, we can also reuse demonstrated grasps. This
is by virtue of the elliptical regions which for similar objects overlap
due to the objects' similar grasp affordances.

The main contributions of our work thus are:
\begin{itemize}
  \item the application of kernel adaptive, mode-hopping MCMC for grasp learning,
  \item active learning of grasps from demonstration without the need for object
    specific knowledge, and
  \item transfer learning of grasps for novel objects given a suitable prior by
    a rough sketch and a few demonstrated grasps of a similar object.
\end{itemize}

We evaluate our proposed learning methods by a series of carefully designed
experiments as presented in Section~\ref{sec:6}. We conclude in Section~\ref{sec:8}
after discussing our experiments in Section~\ref{sec:7}.

\section{Related Work}\label{sec:2}

The majority of research in grasp learning from demonstration builds on recording
hand trajectories~\cite{bohg2014, sahbani2012}. Given such trajectories, Ekvall and
Kragi\'c~\cite{ekvall2004,ekvall2005} present a method that uses Hidden Markov Models
for classification of a demonstrated grasp, whereas Kjellstr{\"o}m et al.~\cite{kjellstrom2008}
and Romero et al.~\cite{romero2008,romero2009}, as well as Aleotti and Caselli~\cite{aleotti2006,aleotti2007}
and Lin and Sun~\cite{lin2014} classify demonstrated grasps by a nearest neighbor
search among already demonstrated grasps. Z\"ollner et al.~\cite{zollner2001} apply
Support Vector Machines for classification of demonstrated grasps.

Instead of classifying the demonstrated grasp type and thus learning concrete grasps
for specific tasks, another idea is to focus on an object's or hand's shape during
demonstration. Li and Pollard~\cite{li2005} introduce a shape-matching algorithm that
consults a database of known hand shapes for suitably grasping an object given its
oriented point representations. Contrary, Kyota et al.~\cite{kyota2005} represent an object
by voxels to identify graspable portions. These portions later are matched against
known poses for suitably grasping an object. Herzog et al.~\cite{herzog2014} learn
gripper 6D poses of grasps which are then generalized to different objects by considering
general shape templates of objects. Ekvall and Kragi\'c~\cite{ekvall2007}, and Tegin et
al.~\cite{tegin2009} extend Ekvall's and Kragi\'c's previous work by considering shape
primitives which are matched to hand shapes for grasping an object. Also, Aleotti and
Caselli~\cite{aleotti2011} extended their work to detect the grasped part of the object,
thus enabling generalization of learned grasps to novel objects. Hsiao and Lozano-P\'erez~\cite{hsiao2006}
segment objects into primitive shapes to map known contact points of grasps to these
shapes. They learn contact points from human demonstration. A key feature
of shape-based learning methods is that they immediately enable transfer learning of
grasps due to the generalization capabilities when only considering the reoccurring parts
of an object's shape.

Yet another approach followed by some researchers is to learn motor skills given
trajectories of human demonstrated grasps. Do et al.~\cite{do2011} interpret a hand as a
spring-mass-damper system, where proper parameterization of this system allows forming grasps.
Kroemer et al.~\cite{kroemer2010} pursue the idea of combining active learning with
reactive control based on vision to learn efficient movement primitives for grasping
from a human demonstrator. Similarly, Pastor et al.~\cite{pastor2011} also consider
the integration of sensory feedback to improve primitive motor skills to learn predictive
models that inherently describe how things should \emph{feel} during execution of a
grasping task.

A more biologically inspired approach is taken by Oztop et al.~\cite{oztop2002} by
employing a neural network resembling the mirror neuron system which is trained by a
human demonstrator for autonomously acquiring grasping skills. Hueser et al.~\cite{hueser2006}
use self-organizing maps to record trajectories which are then used to learn grasping
skills by reinforcement learning.

The work of Granville et al.~\cite{de2006} treats the grasp learning problem from a
probabilistic point of view. Given repeated demonstrations a mixture model for clustering
of grasps is established to eventually learn canonical gripper poses. Faria et al.~\cite{faria2012}
also rely on a series of demonstrations for learning grasps for establishing a probabilistic
model for a grasping task. However, they further incorporate an object centric
volumetric model to infer contact points of grasps, thus also allowing generalizing grasps
to new objects.

Existing research adressing sampling for learning grasps is rather scarce. Detry et
al.~\cite{detry2011} learn grasp affordance densities by establishing an initial
grasp affordance model for an object from early visual cues. This model then is
trained by sampling. Sweeney and Grupen~\cite{sweeney2007} establish a generative
model using an object's visual appearance as well as hand positions and orientations.
Using Gibbs sampling, new grasps then are generated from that model. Kopicki et al.~\cite{kopicki2014}
propose to learn grasps by fitting a gripper's shape to an object's shape
by sampling. Their method allows transfer of grasps by matching the gripper's
shape to shapes of novel objects.

In contrast to existing related work, in this paper we only rely on a gripper's 6D pose
for learning new grasps. Our approach is model-free as we do not rely on an
object model or any object related features. Given a few demonstrated grasps, our
method is capable of learning new grasps for a demonstrated object and transfer
learning of grasps for similar objects.

\section{Background}\label{sec:3}

In what follows we briefly sketch the sampling algorithms our learning methods
build upon.

\subsection{Kernel Adaptive Metropolis Hastings}\label{sec:3.1}

MCMC Kameleon as proposed by Sejdinovic et al.~\cite{sejdinovic2014} is an adaptive
MH sampler approximating highly non-linear target densities $\pi$ in a
reproducing kernel Hilbert space. During
its burn-in phase, at each iteration it obtains a subsample
$\mathbf{z} = \left\{ z_{i} \right\}_{i=1}^{n}$ of the chain history
$\left\{ x_{i} \right\}_{i=0}^{t-1}$ to update the proposal distribution
$q_{\mathbf{z}}(\cdot \mid x)$ by applying kernel PCA on $\mathbf{z}$,
resulting in a low-rank covariance operator $C_{\mathbf{z}}$.
Using $\nu^{2}C_{\mathbf{z}}$ as a covariance (where $\nu$ is a scaling parameter),
a Gaussian measure with mean $k(\cdot,y)$, i.e., $\mathcal{N}(f; k(\cdot,y),
\nu^{2}C_{\mathbf{z}})$, is defined. Samples $f$ from this measure are then
used to obtain target proposals $x^{*}$.

MCMC Kameleon computes pre-images $x^{*} \in \mathcal{X}$ of $f$ by solving the
 non-convex optimization problem
\begin{equation}
  \argmin_{x\in\mathcal{X}}g(x),
\end{equation}
where
\begin{IEEEeqnarray}{rCl}
  g(x) &=& \left\Vert k\left(\cdot,x\right)-f\right\Vert _{\mathcal{H}_{k}}^{2} \\
  &=& k(x,x) - 2k(x,y) - 2 \sum_{i=1}^{n}\mathbf{\beta}_{i}\left [ k(x,z_{i})-\mu_{\mathbf{_z}}(x) \right ], \nonumber
\end{IEEEeqnarray}
$\mu_{\mathbf{z}} = \frac{1}{n}\sum_{i=1}^{n}k(\cdot,z_{i})$,
the empirical measure on $\mathbf{z}$, and $y \in \mathcal{X}$.
Then, by taking a single gradient descent step along the cost function $g(x)$
a new target proposal $x^{*}$ is given by
\begin{equation}
  x^{*} = y - \eta \nabla_{x}g(x)\rvert_{x=y} + \xi
  \label{eqn:MCMC.opt.step}
\end{equation}
where $\mathbf{\beta}$ is a vector of coefficients, $\eta$ is the gradient step size,
and $\xi \sim \mathcal{N}(0,\gamma^{2}I)$ is an additional isotropic exploration term after
the gradient.
 The complete MCMC Kameleon algorithm then is
\begin{itemize}
  \item at iteration $t+1$
  \begin{enumerate}
    \item obtain a subsample $\mathbf{z} = \left\{ z_{i} \right\}_{i=1}^{n}$ of the chain history $\left\{ x_{i} \right\}_{i=0}^{t-1}$,
    \item sample $x^{*} \sim q_{\mathbf{z}}(\cdot \mid x_{t}) = \mathcal{N}(x_{t},\gamma^{2}I + \nu^{2}M_{\mathbf{z},x_{t}}HM_{\mathbf{z},x_{t}}^{T})$,
    \item accept $x^{*}$ with MH acceptance probability $\alpha(x,y) = \min\begin{Bmatrix}
    1, \frac{\pi(y) q(x \mid y)}{\pi(x) q(y \mid x)}
  \end{Bmatrix}$,
  \end{enumerate}
\end{itemize}
where $M_{\mathbf{z},y}=2\eta\left[\nabla_{x}k(x,z_{1})|_{x=y},\ldots,\nabla_{x}k(x,z_{n})|_{x=y}\right]$
is the kernel gradient matrix obtained from the gradient of $g$ at $y$, $\gamma$ is a noise parameter,
and $H$ is an $n \times n$ centering matrix.

\subsection{Generalized Darting Monte-Carlo}\label{sec:3.2}

Generalized Darting Monte Carlo (GDMC)~\cite{smini2011} essentially is
an extension to classic MH samplers by equipping them with mode-hopping capabilities.
Such a mode-hopping behavior is beneficial in case of (i) approximating a highly
non-linear, multimodal target $\pi$, and (ii) counterattack the customary random-walk
behavior of MH samplers by efficiently mixing between modes.

The idea underlying GDMC is to place elliptical jump regions around known modes
of $\pi$. Then, at each iteration, a local MH sampler is interrupted with probability
$P_{check}$, that is, $u_{1} > P_{check}$ where $u_{1} \sim U[0,1]$
to check whether the current state $x_{t}$ is inside a jump region. If $u_{1} < P_{check}$,
sampling continues using the local MH sampler. Otherwise, on $x_{t}$
being inside a jump region, GDMC samples another region to jump to by
\begin{equation}
  P_{i} = \frac{V_{i}}{\sum_{j} V_{j}}
\end{equation}
where $i$ and $j$ are jump region indices. $V$ denotes the n-dimensional elliptical
volume
\begin{equation}
  V = \frac{\pi^\frac{d}{2}\epsilon^{d}\prod_{i=0}^{d}\lambda_{i}}{\Gamma(1 + \frac{d}{2})}
\end{equation}
with $d$ the number of dimensions, $\epsilon$ a scaling factor, and $\lambda_{i}$ the
eigenvalues resulting from the singular value decomposition of the covariance $\Sigma$
of the Markov chain, i.e., $\Sigma = USU^{\top}$ with $S = \text{diag}(\lambda_{i})$.
Observe that $\pi$ in this case denotes the mathematical constant instead of the target
density $\pi$. Given this newly sampled region, GDMC then computes a new state $x_{t+1}$
using the transformation
\begin{equation}
  x_{t+1} = \mu_{x_{t+1}} - U_{x_{t+1}}S_{x_{t+1}}^{\frac{1}{2}}S_{x_{t}}^{-\frac{1}{2}}U_{x_{t}}^{\top}(x_{t} - \mu_{x_{t}})
  \label{eqn:gdmc_sample}
\end{equation}
where $\mu_{\_}$ denotes jump regions' centers (the modes), and $U$ and
$S$ again result from the singular value decomposition of the covariance $\Sigma$
of the Markov chain. GDMC accepts the jump proposal $x_{t+1}$ if $u_{2} > P_{accept}$
where $u_{2} \sim U[0,1]$ and
\begin{equation}
  P_{accept} = min\left [ 1, \frac{n(x_{t})\pi(x_{t})}{n(x_{t+1})\pi(x_{t+1})} \right ]
\end{equation}
with $n(\cdot)$ denoting the number of jump regions that contain a state $x_{i}$.
If $x_{t}$ is outside a jump region, it is counted again, i.e., $x_{t+1} = x_{t}$.

\section{Active Learning of Grasps}\label{sec:4}

We formulate a grasp g as a 7D vector $g = (x, y, z, q_{w}, q_{x}, q_{y}, q_{z})^{\top}$,
where $x, y, z$ denote the cartesian coordinates of
a gripper, and $q_{w}, q_{x}, q_{y}, q_{z}$ its orientation in quaternion notation
about an object. For each grasp, we define a quality measure by the Grasp Wrench Space
(GWS)~\cite{miller1999} denoted $\mu_{GWS}$. This measure then allows us to define
a target density $\pi(g)$ with $g \in \mathcal{X}$. Observe that $\mu_{GWS}$ defines a
valid density function as $\forall g: \mu_{GWS} \geq 0$. Further, by introducing the
normalization constant $Z$ with $Z = \sum_{i=0}^{n} \mu_{GWS}^{i}$ (where $n$ is the
number of known grasps) we have that $\frac{1}{Z}\int\pi(g)\mathrm{d}g = 1$.

Our active learning method takes as an input a rough
sketch of $\pi$ as well as a set of demonstrated grasps. According to Sejdinovic et al.~\cite{sejdinovic2014}
such a rough sketch to initialize MCMC Kameleon does not need to be a proper Markov
chain. Instead, it suffices if it provides good exploratory properties of the target $\pi$.
We construct such a rough sketch by running a purely random walk MH sampler
on the object to be learned. However, we do not take the resulting Markov chain
as an initial sketch but instead the set of proposals generated during the random
walk, irrespective of whether a proposal was accepted or not. The rationale behind this is
that using a purely random MH sampler generally does not result in any learned grasps
(Section~\ref{sec:7}). Hence, the resulting Markov chain essentially would not contain
any samples and thus does not inhibit good exploratory properties of $\pi$. On the other
hand, the set of proposals as generated during the random walk encapsulates enough
information regarding an approximation of the shape $\pi$. Thence, it suffices as a rough
sketch to initialize MCMC Kameleon. The random walk MH sampler employed for this
uses a Gaussian proposal for the position and a von-Mises-Fisher proposal for the
orientation, i.e.,
\begin{equation*}
  \begin{split}
    &g^{*}_{\mathrm{pos}} = \mathcal{N}(g_{\mathrm{pos}}^{t}, \Sigma)\\
    &g^{*}_{\mathrm{ori}} = \mathcal{C}_{4}(\kappa)\exp(\kappa g_{\mathrm{ori}}^{t\;\ \top} \mathbf{x}),
  \end{split}
\end{equation*}
where $\kappa$ is a concentration parameter and $\mathbf{x}$ a p-dimensional
unit direction vector. We use the same probability measures as defined
for MCMC Kameleon by the GWS\@.

In a real-world environment, the set of demonstrated grasps would be established
by moving the robot's gripper towards a position and into a pose, where it can grasp
the object. The gripper's position in cartesian space as well as its
orientation about the object in SO(3) are then recorded and treated as
a demonstrated grasp. In this work however we only study our grasp learning
methods in simulation (Section~\ref{sec:6}). Thus, we randomly select points on
the object's surface to then find a grasp by optimizing the gripper's pose about
its orientation~\cite{zech2015}.

Given a rough sketch of $\pi$ and a set of user demonstrated grasps, the complete
learning method then can be sketched as:
\begin{itemize}
  \item at iteration $t+1$
  \begin{enumerate}
    \item[-] attempt to perform a jump move according to the procedure as outlined in
      Section~\ref{sec:3.2},
    \item[-] otherwise, sample locally using MCMC Kameleon as outlined in
      Section~\ref{sec:3.1}.
  \end{enumerate}
\end{itemize}
As a kernel $k$ for MCMC Kameleon we chose a Gaussian kernel. Whilst not rigorously
applicable in quaternion space, it allows us to model the dependency between a gripper's
position and its orientation. Further, during our experiments we found that a Gaussian
kernel works quite well in practice.

\section{Transfer Learning of Grasps}\label{sec:5}

Transfer learning fundamentally captures the idea of reusing existing knowledge or
already acquired skills to solve problems similar to the original one. For transfer
learning of grasps for novel, as of yet unseen objects this ultimately boils down
to reusing both the Markov chains constructed when learning to grasp a known object
and the respective set of user demonstrated grasps. A crucial factor for the
success of this procedure however is that the known and the novel object are similar
in \emph{shape} and \emph{size} (e.g., a plate and a soup plate). Given that this
constraint is satisfied reusing of Markov chains and grasps is feasible due to
both MCMC Kameleon's learning behavior during a burn-in phase, as well as GDMC's
construction of elliptical regions around known modes. As discussed in Section~\ref{sec:3.2},
GDMC samples a new state $x_{t+1}$ by applying the transformation as outlined in
equation~\eqref{eqn:gdmc_sample}. As this transformation does not
tie a new state $x_{t+1}$ exactly to a mode, but instead into the elliptical
region constructed around it, there is a high probability that a
new state $x_{t+1}$ is close to a mode of the grasp density $\pi$ for the novel object.
Thus, jump moves as done by GDMC are valid in the sense that they again nudge the
proposal generating process close to modes of $\pi$. Apart from that, recycling of
existing Markov chains and already demonstrated grasps yields substantial time
savings by sidestepping both construction of a rough sketch for a novel
object and by having a user demonstrate new grasps.

\section{Experimental Method}\label{sec:6}

\begin{figure}
  \centering
  \resizebox{\columnwidth}{!}{
    \includegraphics{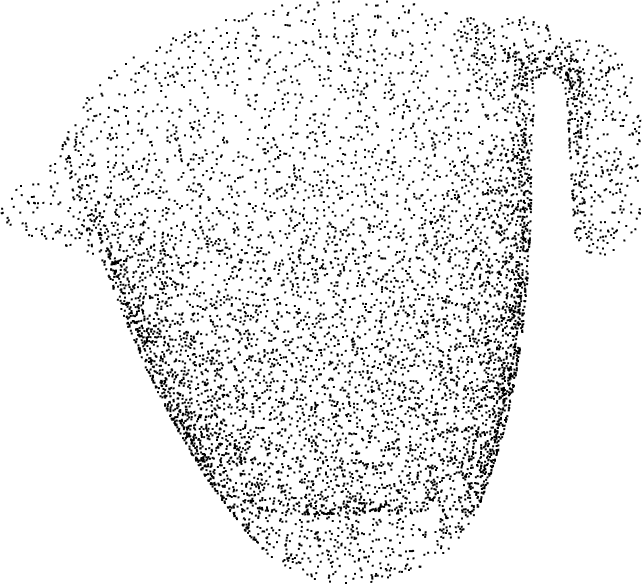}\hspace{1cm}
    \includegraphics{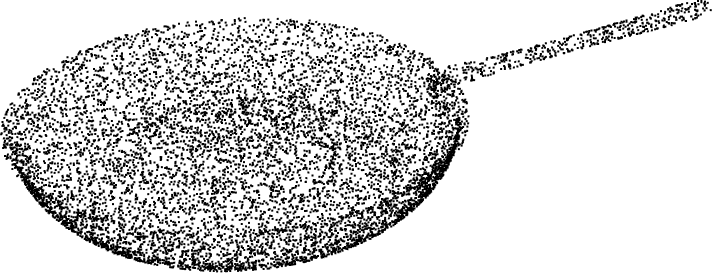}
    \includegraphics{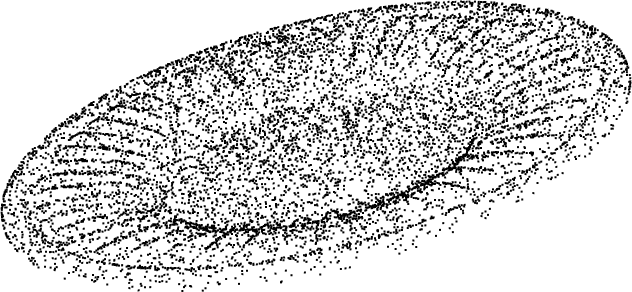}}
  \caption{Object set used for grasp learning.}
  \label{fig:object_set_al}
\end{figure}

\begin{figure}
  \centering
  \resizebox{\columnwidth}{!}{
    \includegraphics{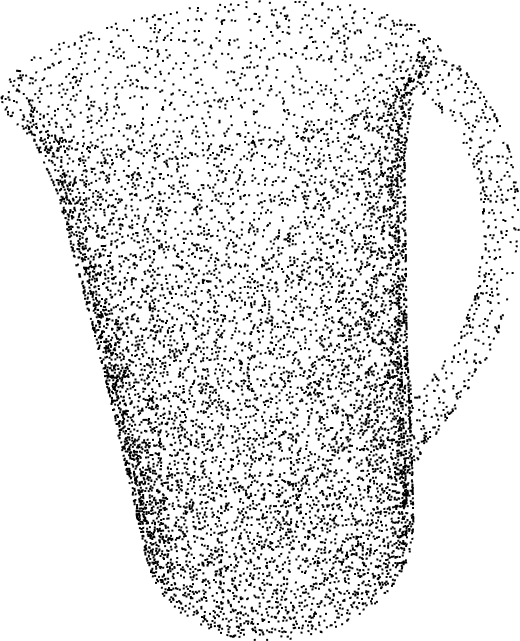}\hspace{1cm}
    \includegraphics{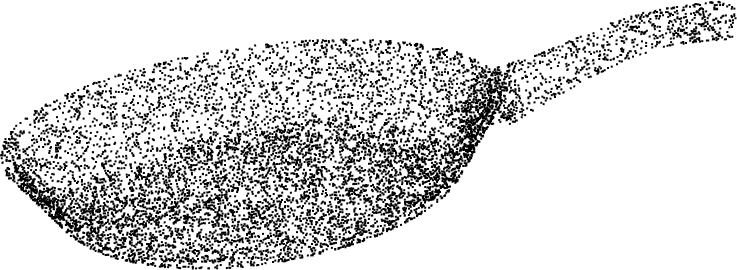}
    \includegraphics{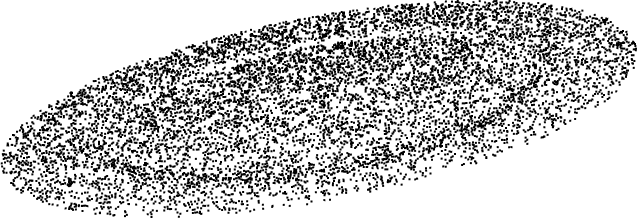}}\vspace{.5cm}
  \resizebox{\columnwidth}{!}{
    \includegraphics{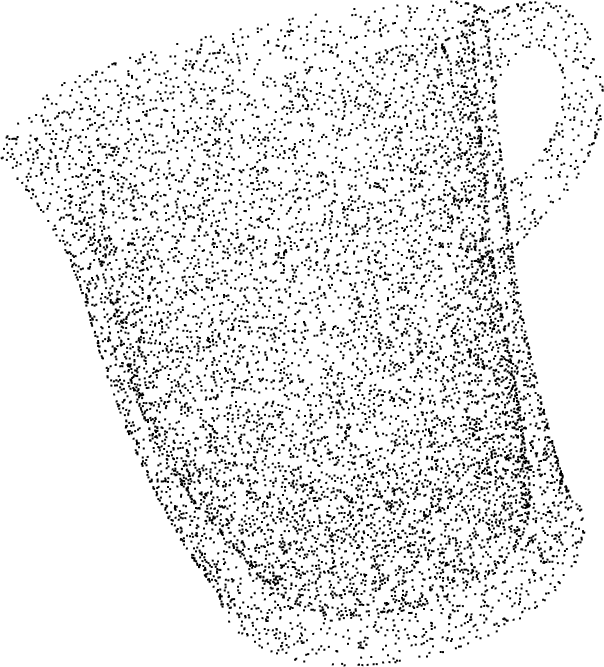}\hspace{1cm}
    \includegraphics{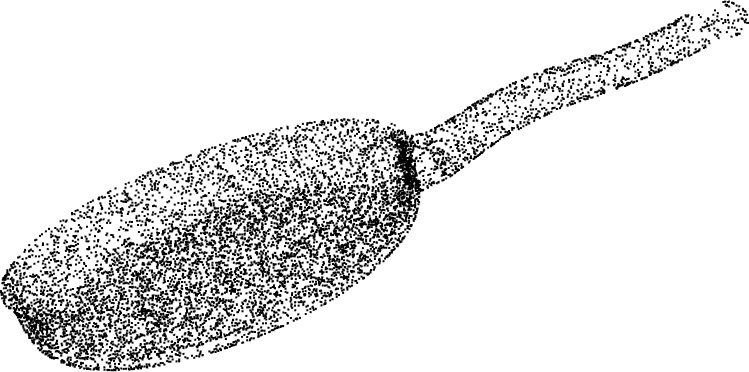}
    \includegraphics{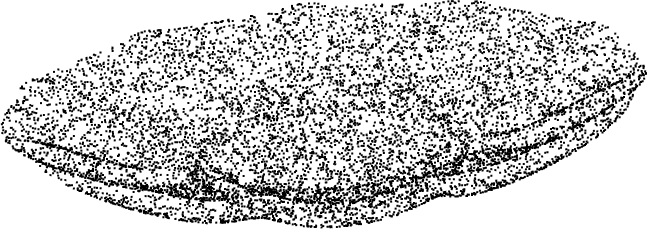}}
  \caption{Object set used for transfer learning.}
  \label{fig:object_set_tl}
\end{figure}

We evaluated our learning methods with 9 different objects as depicted in
Figures~\ref{fig:object_set_al} and~\ref{fig:object_set_tl}. In total we
performed 5 experiments using RobWork~\cite{ellekilde2010}, a robotics
and grasp simulator. Our first experiment acts as a baseline that allows us
to compare the efficiency of our active learning method to a purely random
walk (as sketched in Section~\ref{sec:4}). The next two experiments were designed
to evaluate our active learning method. First, MCMC Kameleon was initialized
with a random sketch, that is, a randomly generated set of gripper poses
essentially capturing no properties of $\pi$. Secondly, MCMC Kameleon was
initialized with a nonrandom sketch consisting of the trace of a purely
random walk MH sampler as discussed in Section~\ref{sec:4}.

The last two experiments were designed to evaluate our transfer learning method.
For initializing MCMC Kameleon for both of these we reused the Markov chains
constructed when learning to grasp a similar object. As necessary user demonstrated grasps,
in the penultimate experiment we used similar modes, that is, grasps that were demonstrated
for a similar object. For the last experiment we used grasps demonstrated on the actual
object. This choice of experimental design allows us to evaluate whether our proposed
transfer learning method can work with no object specific knowledge at all.

Table~\ref{tab:params} shows our parameterization of MCMC Kameleon
and GDMC for our experiments. The values were established during
a series of preliminary experiments.
\begin{table}
  \renewcommand{\arraystretch}{1.8}
  \centering
  \caption{Parameterization of MCMC Kameleon and GDMC for our experiments.}
  \begin{tabular}{@{}ccccccc@{}}
    Iterations & $\gamma$ & Subsample size& $\nu$~\cite{sejdinovic2014} &  Burn-in & $P_{check}$ & $\epsilon$ \\\hline
    1000       & 0.0001   & 100           & $\frac{2.38}{\sqrt{6}}$ &  100     & 0.6         & 0.7
  \end{tabular}
  \label{tab:params}
\end{table}
For all experiments we used 5 demonstrated grasps.

\section{Results and Discussion}\label{sec:7} 

\begin{figure}
  \centering
    \resizebox{\columnwidth}{!}{%
    \includegraphics{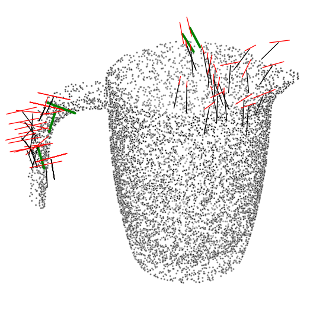}%
    \includegraphics{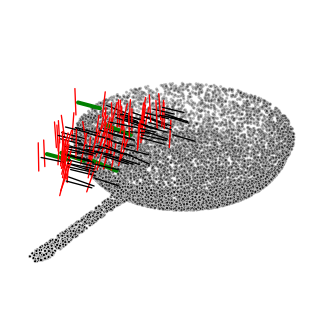}%
    \includegraphics{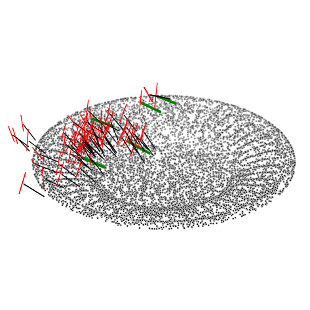}}
    \resizebox{\columnwidth}{!}{%
    \includegraphics{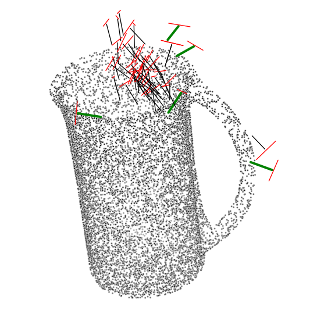}%
    \includegraphics{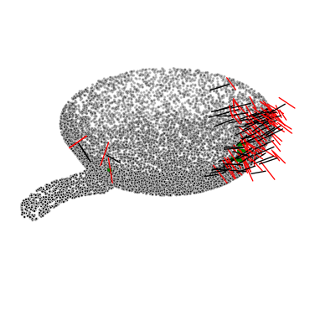}%
    \includegraphics{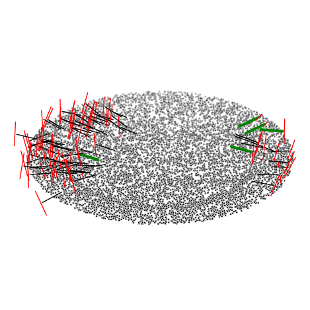}}
    \resizebox{\columnwidth}{!}{%
    \includegraphics{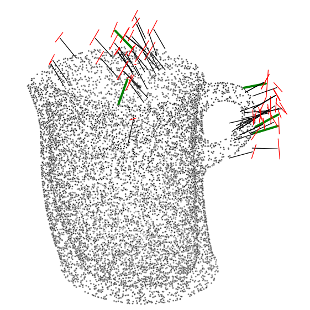}%
    \includegraphics{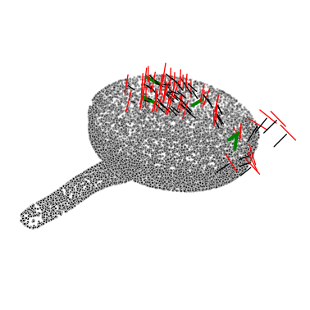}%
    \includegraphics{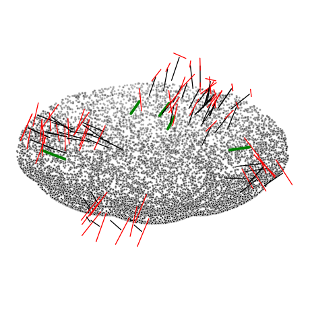}}
  \caption{Results for learning grasps for the objects from Figure~\ref{fig:object_set_al}
    (top row) and for transfer learning of grasps for corresponding
    objects from Figure~\ref{fig:object_set_tl} (middle and bottom rows).
    Observe that grasps are rather unevenly distributed; this results from
    using only $100+1000$ iterations. Black lines denote the orientation of the gripper,
    red lines its span; demonstrated grasps are colored green.}
  \label{fig:results}
\end{figure}

For all three objects from Figure~\ref{fig:object_set_al} our active learning method
found an additional number of grasps as is shown in Table~\ref{tab:res_al}. Further,
Table~\ref{tab:res_al} clearly shows that combining MCMC
Kameleon with GDMC drastically outperforms a purely random walk. Also, our active
learning method actually works without any knowledge except a few user demonstrated
grasps. This is visible from Table~\ref{tab:res_al} when we did our experiments with
MCMC Kameleon initialized with a random sketch. Also visible from Table~\ref{tab:res_al},
the more complex an object's shape, the more difficult it is to learn grasps for it (cf.~the
pitcher with the pan or plate; generally, for the former, fewer grasps were learned). We thus
infer that our active learning method for grasping from user demonstration is successful.
The top row from Figure~\ref{fig:results} shows grasps resulting from our active learning
method when applied to the objects from Figure~\ref{fig:object_set_al}.
\begin{table*}
  \renewcommand{\arraystretch}{1.8}
  \centering
  \caption{Results for active learning of grasps (succ. = success, slip. = slipped, coll. = collision).}
  \begin{tabular}{@{}lcccc|cccc|cccc@{}}
             & \multicolumn{4}{c|}{Random Walk} & \multicolumn{8}{c}{MCMC Kameleon combined with GDMC} \\
             & \multicolumn{4}{c|}{} & \multicolumn{4}{c|}{Random Initialization} & \multicolumn{4}{c}{Biased Initialization} \\\hline
             & Succ. & Slip. & Coll. & Miss & Succ. & Slip. & Coll. & Miss & Succ. & Slip. & Coll. & Miss \\\hline
     Pitcher & 0 & 3  & 433 & 664 & 37 & 48  & 536 & 479 & 49 & 80 & 661 & 310  \\
     Pan     & 2 & 10 & 377 & 711 & 39 & 50  & 418 & 593 & 66 & 54 & 477 & 503  \\
     Plate   & 1 & 28 & 361 & 710 & 43 & 146 & 679 & 232 & 59 & 91 & 662 & 288
  \end{tabular}
  \label{tab:res_al}
\end{table*}

For transfer learning of grasps for novel, as of yet unseen objects we
arrive at the same conclusion as for active learning of grasps. Our
learning method again was successful in finding grasps (Table~\ref{tab:res_tl}).
Further, as is evident from Table~\ref{tab:res_tl} our learning method generally
is able to learn new grasps for novel objects without the need for any user demonstrated
grasp for the specific object (cf.~pans and plates). However, as can also be
seen from the data in Table~\ref{tab:res_tl} our transfer learning method may
fail drastically. For both pitchers our learning method failed in learning grasps
using similar modes. This is by virtue of the vastly differing sizes and geometries of
the pitchers. Obviously, taking the modes and the Markov chain of the pitcher from Figure~\ref{fig:object_set_al}
as a rough sketch as well as initial modes for the grasp density of the tall pitcher
from Figure~\ref{fig:object_set_tl} (top row) is a lead balloon. The discrepancy of
the size and the geometry of these objects is just too big. The bottom rows from Figure~\ref{fig:results} show the outcomes
of our transfer learning method when applied to the objects from Figure~\ref{fig:object_set_tl}.
\begin{table}
  \renewcommand{\arraystretch}{1.8}
  \centering
  \caption{Results for transfer learning of grasps. The upper block corresponds
    to the top row of Figure~\ref{fig:object_set_tl}; the lower block to the bottom
    row (succ. = success, slip. = slipped, coll. = collision).}
  \begin{tabular}{@{}lcccc|cccc@{}}
             & \multicolumn{4}{c|}{Modes of a similar object} & \multicolumn{4}{c}{Modes of the actual object} \\\hline
             & Succ. & Slip. & Coll. & Miss & Succ. & Slip. & Coll. & Miss     \\\hline
     Pitcher & 0 & 700  & 400 & 0 & 42 & 109 & 576 & 373 \\
     Pan     & 54 & 43 & 679 & 324 & 66 & 90 & 787 & 157 \\
     Plate   & 66 & 107 & 633 & 294 & 69 & 164 & 755 & 112 \\\hline
     Pitcher & 0 & 154  & 946 & 0 & 63 & 130 & 487 & 420 \\
     Pan     & 38 & 46 & 716 & 300 & 52 & 73 & 755 & 220 \\
     Plate   & 60 & 67 & 730 & 243 & 63 & 86 & 771 & 180
  \end{tabular}
  \label{tab:res_tl}
\end{table}

To conclude, we state that the combination of MCMC Kameleon and GDMC yields
good exploratory properties when searching for feasible grasps in an object's
grasp space by requiring no more input than a few user demonstrated grasps as
6D gripper poses. This is evident from both Tables~\ref{tab:res_al} and~\ref{tab:res_tl}
in that the number of misses generally is substantially smaller than the total
number of collisions and grasps where the object slipped out of the gripper.

\section{Conclusions}\label{sec:8}

We have presented both a novel method for active learning of grasps
as well as a novel method for transfer learning of grasps, suitably biased by prior
experience. We have shown that learning of grasps is feasible without the
requirement of object related knowledge. Our learning methods require nothing more
than a few demonstrated grasps.

Both our learning methods are grounded on MCMC sampling, more specifically
a combination of MCMC Kameleon and GDMC\@. These algorithms each have advantageous
characteristics. MCMC Kameleon allows sampling from highly non-linear distributions,
whereas GDMC tackles the issue of properly exploring a multimodal distribution. We
found that a combination of both ideally fits the problem of active and transfer
learning of grasps. Our results as shown in Tables~\ref{tab:res_al} and~\ref{tab:res_tl}
further undermine our conclusions.

Concerning transfer learning of grasp for novel, as of yet unseen objects, we
further want to highlight two observations. First, reusing an existing Markov chain
allows boosting of our learning methods by avoiding construction of an initial
rough sketch of $\pi$ for an object. Secondly, given that two objects are (i) not
too dissimilar in shape and size, and (ii) properly aligned by the same canonical pose,
then our transfer learning method is capable of learning grasps for novel objects
without any object specific knowledge.
\balance

\bibliographystyle{IEEEtran}
\bibliography{references}

\begin{thebibliography}{10}
\providecommand{\url}[1]{#1}
\csname url@samestyle\endcsname
\providecommand{\newblock}{\relax}
\providecommand{\bibinfo}[2]{#2}
\providecommand{\BIBentrySTDinterwordspacing}{\spaceskip=0pt\relax}
\providecommand{\BIBentryALTinterwordstretchfactor}{4}
\providecommand{\BIBentryALTinterwordspacing}{\spaceskip=\fontdimen2\font plus
\BIBentryALTinterwordstretchfactor\fontdimen3\font minus
  \fontdimen4\font\relax}
\providecommand{\BIBforeignlanguage}[2]{{%
\expandafter\ifx\csname l@#1\endcsname\relax
\typeout{** WARNING: IEEEtran.bst: No hyphenation pattern has been}%
\typeout{** loaded for the language `#1'. Using the pattern for}%
\typeout{** the default language instead.}%
\else
\language=\csname l@#1\endcsname
\fi
#2}}
\providecommand{\BIBdecl}{\relax}
\BIBdecl

\bibitem{bohg2014}
J.~Bohg, A.~Morales, T.~Asfour, and D.~Kragic, ``{Data-driven Grasp Synthesis
  --- A Survey},'' \emph{Robotics, IEEE Transactions on}, vol.~30, no.~2, pp.
  289--309, 2014.

\bibitem{sahbani2012}
A.~Sahbani, S.~El-Khoury, and P.~Bidaud, ``{An Overview of 3D Object Grasp
  Synthesis Algorithms},'' \emph{Robotics and Autonomous Systems}, vol.~60,
  no.~3, pp. 326--336, 2012.

\bibitem{bala2012}
R.~Balasubramanian, L.~Xu, P.~D. Brook, J.~R. Smith, and Y.~Matsuoka,
  ``{Physical Human Interactive Guidance: Identifying Grasping Principles from
  Human-planned Grasps},'' \emph{IEEE Transactions on Robotics}, vol.~28,
  no.~4, pp. 899--910, 2012.

\bibitem{hastings1970}
W.~K. Hastings, ``{Monte Carlo Sampling Methods Using Markov Chains and Their
  Applications},'' \emph{Biometrika}, vol.~57, no.~1, pp. 97--109, 1970.

\bibitem{sejdinovic2014}
D.~Sejdinovic, H.~Strathmann, M.~L. Garcia, C.~Andrieu, and A.~Gretton,
  ``{Kernel Adaptive Metropolis-Hastings},'' in \emph{Proceedings of the 31st
  International Conference on Machine Learning}, E.~P. Xing and T.~Jebara,
  Eds., vol.~32, 2014, pp. 1665--1673.

\bibitem{smini2011}
C.~Sminchisescu and M.~Welling, ``{Generalized darting Monte Carlo},''
  \emph{Pattern Recognition}, vol.~44, no. 10–11, pp. 2738--2748, 2011.

\bibitem{ekvall2004}
S.~Ekvall and D.~Kragi{\'c}, ``{Interactive Grasp Learning Based on Human
  Demonstration},'' in \emph{IEEE International Conference on Robotics and
  Automation, ICRA'04}, vol.~4.\hskip 1em plus 0.5em minus 0.4em\relax IEEE,
  2004, pp. 3519--3524.

\bibitem{ekvall2005}
------, ``{Grasp Recognition for Programming by Demonstration},'' in \emph{IEEE
  International Conference on Robotics and Automation, ICRA'05.}\hskip 1em plus
  0.5em minus 0.4em\relax IEEE, 2005, pp. 748--753.

\bibitem{kjellstrom2008}
H.~Kjellstr{\"o}m, J.~Romero, and D.~Kragi{\'c}, ``{Visual Recognition of
  Grasps for Human-to-Robot Mapping},'' in \emph{IEEE/RSJ International
  Conference on Intelligent Robots and Systems, IROS'08.}\hskip 1em plus 0.5em
  minus 0.4em\relax IEEE, 2008, pp. 3192--3199.

\bibitem{romero2008}
J.~Romero, H.~Kjellstr{\"o}m, and D.~Kragi{\'c}, ``{Human-to-Robot Mapping of
  Grasps},'' in \emph{IEEE/RSJ International Conference on Intelligent Robots
  and Systems, Workshop on Grasp and Task Learning by Imitation, IROS'08},
  2008.

\bibitem{romero2009}
------, ``{Modeling and Evaluation of Human-to-Robot Mapping of Grasps},'' in
  \emph{International Conference on Advanced Robotics, ICAR'09.}\hskip 1em plus
  0.5em minus 0.4em\relax IEEE, 2009, pp. 1--6.

\bibitem{aleotti2006}
J.~Aleotti and S.~Caselli, ``{Grasp Recognition in Virtual Reality for Robot
  Pregrasp Planning by Demonstration},'' in \emph{IEEE International Conference
  on Robotics and Automation, ICRA'06.}\hskip 1em plus 0.5em minus 0.4em\relax
  IEEE, 2006, pp. 2801--2806.

\bibitem{aleotti2007}
------, ``{Robot Grasp Synthesis from Virtual Demonstration and
  Topology-preserving Environment Reconstruction},'' in \emph{IEEE/RSJ
  International Conference on Intelligent Robots and Systems, IROS'07.}\hskip
  1em plus 0.5em minus 0.4em\relax IEEE, 2007, pp. 2692--2697.

\bibitem{lin2014}
Y.~Lin and Y.~Sun, ``{Robot Grasp Planning Based on Demonstrated Grasp
  Strategies},'' \emph{The International Journal of Robotics Research}, pp.
  26--42, 2014.

\bibitem{zollner2001}
R.~Z{\"o}llner, O.~Rogalla, R.~Dillmann, and J.~Zoellner, ``{Dynamic Grasp
  Recognition Within the Framework of Programming by Demonstration},'' in
  \emph{IEEE International Workshop on Robot and Human Interactive
  Communication.}\hskip 1em plus 0.5em minus 0.4em\relax IEEE, 2001, pp.
  418--423.

\bibitem{li2005}
Y.~Li and N.~S. Pollard, ``{A Shape Matching Algorithm for Synthesizing
  Humanlike Enveloping Grasps},'' in \emph{IEEE-RAS International Conference on
  Humanoid Robots, HUMANOIDS'05.}\hskip 1em plus 0.5em minus 0.4em\relax IEEE,
  2005, pp. 442--449.

\bibitem{kyota2005}
F.~Kyota, T.~Watabe, S.~Saito, and M.~Nakajima, ``{Detection and Evaluation of
  Grasping Positions for Autonomous Agents},'' in \emph{International
  Conference on Cyberworlds.}\hskip 1em plus 0.5em minus 0.4em\relax IEEE,
  2005, pp. 460--468.

\bibitem{herzog2014}
A.~Herzog, P.~Pastor, M.~Kalakrishnan, L.~Righetti, J.~Bohg, T.~Asfour, and
  S.~Schaal, ``{Learning of Grasp Selection Based on Shape-templates},''
  \emph{Autonomous Robots}, vol.~36, no. 1-2, pp. 51--65, 2014.

\bibitem{ekvall2007}
S.~Ekvall and D.~Kragi{\'c}, ``{Learning and Evaluation of the Approach Vector
  for Automatic Grasp Generation and Planning},'' in \emph{IEEE International
  Conference on Robotics and Automation, ICRA'07.}\hskip 1em plus 0.5em minus
  0.4em\relax IEEE, 2007, pp. 4715--4720.

\bibitem{tegin2009}
J.~Tegin, S.~Ekvall, D.~Kragic, J.~Wikander, and B.~Iliev,
  ``{Demonstration-based Learning and Control for Automatic Grasping},''
  \emph{Intelligent Service Robotics}, vol.~2, no.~1, pp. 23--30, 2009.

\bibitem{aleotti2011}
J.~Aleotti and S.~Caselli, ``{Part-based Robot Grasp Planning from Human
  Demonstration},'' in \emph{IEEE International Conference on Robotics and
  Automation, ICRA'11}.\hskip 1em plus 0.5em minus 0.4em\relax IEEE, 2011, pp.
  4554--4560.

\bibitem{hsiao2006}
K.~Hsiao and T.~Lozano-Perez, ``{Imitation Learning of Whole-body Grasps},'' in
  \emph{IEEE/RSJ International Conference on Intelligent Robots and Systems,
  IROS'06.}\hskip 1em plus 0.5em minus 0.4em\relax IEEE, 2006, pp. 5657--5662.

\bibitem{do2011}
M.~Do, T.~Asfour, and R.~Dillmann, ``{Towards a Unifying Grasp Representation
  for Imitation Learning on Humanoid Robots},'' in \emph{IEEE International
  Conference on Robotics and Automation, ICRA'11.}\hskip 1em plus 0.5em minus
  0.4em\relax IEEE, 2011, pp. 482--488.

\bibitem{kroemer2010}
O.~Kroemer, R.~Detry, J.~Piater, and J.~Peters, ``{Combining Active Learning
  and Reactive Control for Robot Grasping},'' \emph{Robotics and Autonomous
  Systems}, vol.~58, no.~9, pp. 1105--1116, 2010.

\bibitem{pastor2011}
P.~Pastor, L.~Righetti, M.~Kalakrishnan, and S.~Schaal, ``{Online Movement
  Adaptation based on Previous Sensor experiences},'' in \emph{IEEE/RSJ
  International Conference on Intelligent Robots and Systems, IROS'11.}\hskip
  1em plus 0.5em minus 0.4em\relax IEEE, 2011, pp. 365--371.

\bibitem{oztop2002}
E.~Oztop and M.~A. Arbib, ``{Schema Design and Implementation of the
  Grasp-related Mirror Neuron System},'' \emph{Biological Cybernetics},
  vol.~87, no.~2, pp. 116--140, 2002.

\bibitem{hueser2006}
M.~Hueser, T.~Baier, and J.~Zhang, ``{Learning of Demonstrated Grasping Skills
  by Stereoscopic Tracking of Human Head Configuration},'' in \emph{IEEE
  International Conference on Robotics and Automation, ICRA'06}.\hskip 1em plus
  0.5em minus 0.4em\relax IEEE, 2006, pp. 2795--2800.

\bibitem{de2006}
C.~de~Granville, J.~Southerland, and A.~H. Fagg, ``{Learning Grasp Affordances
  Through Human Demonstration},'' in \emph{International Conference on
  Development and Learning, ICDL’06.}\hskip 1em plus 0.5em minus 0.4em\relax
  IEEE, 2006.

\bibitem{faria2012}
D.~R. Faria, R.~Martins, J.~Lobo, and J.~Dias, ``{Extracting Data from Human
  Manipulation of Objects Towards Improving Autonomous Robotic Grasping},''
  \emph{Robotics and Autonomous Systems}, vol.~60, no.~3, pp. 396--410, 2012.

\bibitem{detry2011}
R.~Detry, D.~Kraft, O.~Kroemer, L.~Bodenhagen, J.~Peters, N.~Kr{\"u}ger, and
  J.~Piater, ``{Learning Grasp Affordance Densities},'' \emph{Paladyn}, vol.~2,
  no.~1, pp. 1--17, 2011.

\bibitem{sweeney2007}
J.~D. Sweeney and R.~Grupen, ``{A Model of Shared Grasp Affordances From
  Demonstration},'' in \emph{IEEE-RAS International Conference on Humanoids
  Robots, HUMANOIDS'07.}\hskip 1em plus 0.5em minus 0.4em\relax IEEE, 2007, pp.
  27--35.

\bibitem{kopicki2014}
M.~Kopicki, R.~Detry, F.~Schmidt, C.~Borst, R.~Stolkin, and J.~Wyatt,
  ``{Learning Dexterous Grasps that Generalise to Novel Objects by Combining
  Hand and Contact Models},'' in \emph{IEEE International Conference on
  Robotics and Automation, ICRA'14.}, May 2014, pp. 5358--5365.

\bibitem{miller1999}
A.~T. Miller and P.~K. Allen, ``{Examples of 3D Grasp Quality Computations},''
  in \emph{IEEE International Conference on Robotics and Automation, 1999.
  Proceedings, ICRA'99.}, vol.~2.\hskip 1em plus 0.5em minus 0.4em\relax IEEE,
  1999, pp. 1240--1246.

\bibitem{zech2015}
P.~Zech, H.~Xiong, and J.~Piater, ``{Rotation Optimization on the Unit
  Quaternion Manifold and its Application for Robotic Grasping},'' in \emph{IMA
  Conference on Mathematics of Robotics}.\hskip 1em plus 0.5em minus
  0.4em\relax IMA, 2015, to appear.

\bibitem{ellekilde2010}
L.-P. Ellekilde and J.~A. Jorgensen, ``{RobWork: A Flexible Toolbox for
  Robotics Research and Education},'' in \emph{41st International Symposium on
  Robotics (ISR) and 2010 6th German Conference on Robotics (ROBOTIK)}.\hskip
  1em plus 0.5em minus 0.4em\relax VDE, 2010, pp. 1--7.

\end{thebibliography}

\end{document}